\documentclass[10pt, a4paper, onecolumn]{article}

\usepackage[numbers]{natbib}

\usepackage[T1]{fontenc}
\usepackage{lmodern}

\usepackage{natbib}
\usepackage[utf8]{inputenc} 
\usepackage{booktabs}       
\usepackage{amsfonts}       
\usepackage{nicefrac}       
\usepackage{microtype}      
\usepackage{xcolor}         
\usepackage{graphicx}
\usepackage{amsmath,amssymb,amsfonts}
\usepackage{bm}
\usepackage{url}

\DeclareMathOperator*{\argmin}{arg\,min}
\usepackage{array,multirow,graphicx}

\usepackage{geometry}
\newgeometry{vmargin={15mm}, hmargin={12mm,17mm}} 

\title{A temporally quantized distribution of pupil diameters as a new feature for cognitive load classification}

\author{
	Wolfgang Fuhl
	\and
	Susanne Zabel
	\and
	Theresa Harbig
	\and
	Julia Astrid Moldt
	\and
	Teresa Festl Wiete
	\and
	Anne Herrmann Werner
	\and
	Kay Nieselt
	}

\date{}

\begin{document}
	
	\maketitle
	
	\begin{abstract}
		In this paper, we present a new feature that can be used to classify cognitive load based on pupil information. The feature consists of a temporal segmentation of the eye tracking recordings. For each segment of the temporal partition, a probability distribution of pupil size is computed and stored. These probability distributions can then be used to classify the cognitive load. The presented feature significantly improves the classification accuracy of the cognitive load compared to other statistical values obtained from eye tracking data, which represent the state of the art in this field. The applications of determining Cognitive Load from pupil data are numerous and could lead, for example, to pre-warning systems for burnouts. \\
		Link: \url{https://es-cloud.cs.uni-tuebingen.de/d/8e2ab8c3fdd444e1a135/?p=%2FCognitiveLoadFeature&mode=list}
	\end{abstract}

	\section{Introduction}
	It is well known, that cognitive load or workload can be measured via the pupil dilation~\cite{beatty1982task,WF042019,zekveld2011cognitive}. The main challenge of extracting information from the pupil dilation is the effect of changing lightning conditions on the pupil size, which are common in real world scenarios. Since the effect on the pupil size of different luminous intensities is a multiple compare to the effect based on cognitive load, studies in this area have to be conducted under strict lightning conditions~\cite{beatty1982task,zekveld2011cognitive,WTCKWE092015,WTTE032016,appel2018cross,abdrabou2021think}. The effect of cognitive load on the pupil size scales linear with the task difficulty as it was shown in ~\cite{beatty1982task}. Based on these findings, many studies where conducted in different possible application areas like driving~\cite{kun2013feasibility,marquart2015review,palinko2012exploring}, aviation~\cite{peysakhovich2015pupil,biondi2023distracted}, medicine~\cite{szulewski2015use,vella2021exploratory}, and psychology~\cite{klingner2011effects,laeng2011pupillary}. In contrast to usually used questionnaires, the pupil dilation is an objective measure, and therefore more reliable. In addition, the pupil dilation could be used in real applications since the user would not have to fill out a questionnaire, if the problem with the changing lightning conditions is solved. An alternative to the pupil dilation also exists with the electroencephalogram (EEG)~\cite{swerdloff2021dry,antonenko2010using}. The main disadvantage of the EEG is the large amount of time required to properly attach it and confirm the signal quality ~\cite{swerdloff2021dry,friedman2019eeg,kumar2016measurement}. With the development of dry EEG electrodes this disadvantage could be circumvented but so far the signal quality of dry EEG electrodes is not sufficient ~\cite{swerdloff2021dry,shad2020impedance,raduntz2018signal,fiedler2010novel}. Therefore, the pupil dilation is still a very interesting source for cognitive load~\cite{VECETRA2020,CORR2017FuhlW1,ETRA2018FuhlW}. 
	
	Modern studies try to improve the cognitive load estimation with various additional eye tracking statistics, like fixation duration, saccade velocity and the blink rate or other sources like skin conductivity or heart rate~\cite{appel2018cross,gowrisankaran2012asthenopia,C2019,FFAO2019,fuhl2022hpcgen,krejtz2018eye,wilson2021objective,fietz2022pupillometry}. While the blink rate was shown to contain information about the cognitive load~\cite{gowrisankaran2012asthenopia}, the other statistics seam to contain to much experiment specific information and therefore falsify the result ~\cite{appel2018cross,ktistakis2022colet}. In addition, most of the reported results published are still not satisfying especially if it comes to cross subject evaluations~\cite{appel2018cross}.
	
	While other approaches tried to tackle the problem by finding better features (like micro saccades~\cite{krejtz2018eye,MEMD2021FUHL}) and feature combinations our approach steams from the machine learning perspective. We propose a novel feature which separates the data stream into separate distributions which only contain the changing pupil size information. In addition, we used a public data set so our results can be reproduced and compared to other approaches.


	\section{Related Work}
	There exist three types of measures for the cognitive load, namely subjective, physiological and performance measures [28]. In case of the subjective measure, a questionnaire or an interview with the subject is evaluated. For this type of measure, usually the NASA-TLX questionnaire is used ~\cite{hart1988development}. Disadvantages of this type of measure are that they cannot be used to detect rapid changes, nor can they be used for automatic assessment and usage in automotive cars, for example. The physiological measures are sensor responses which correlate with the cognitive load. Those measures are pupil dilation, heart-rate variability, skin conductivity, EEG, etc. ~\cite{beatty2000pupillary,ikehara2005assessing,iqbal2005towards,hess1964pupil,van2018pupil,swerdloff2021dry}. Since all these values are measured by sensors, they are subject to certain errors and do not change only based on the cognitive load. Other dependencies of The physiological measures are anxiety ~\cite{chen2009cognitive}, arousal ~\cite{kassem2017diva}, physical activity ~\cite{romance2018cognitive}, stress ~\cite{conway2013effect}, fatigue ~\cite{tanaka2015effects}, and light ~\cite{reeves1920response}. Since the pupil dilation is a possible candidate for real-time usage, many researchers investigated this feature ~\cite{chen2014using,chen2013automatic,hussain2011classification,kiefer2016measuring}. In \cite{ktistakis2022colet} fixation and saccade statistics are proposed. Those features bear the disadvantage that they could be experiment specific and not relevant for cognitive load. Therefore, those features could falsify the results. Micro saccades are proposed in \cite{krejtz2018eye} but have the disadvantage of a very strict setup, where the user is not allowed to move his eyes. Additionally, the eye tracker must be very accurate to detect micro saccades.
	The last measure is based on the performance. Here, the evaluation uses the variation in human performance based on different levels of cognitive load. In case of a cognitive overload the performance should decrease~\cite{yerkes1908relation,veltman2005role,babiloni2019mental,sevcenko2021measuring}. While this is the most intuitive measure and simple to apply, it still has some limiting factors. One problem is that the performance can drop due to multiple influencing factors like motivation or arousal~\cite{brunken2010measuring}. This means that the cognitive load cannot be analyzed independently of the other factors. In addition, this approach is also not online applicable~\cite{sevcenko2021measuring}.

	\section{Method}

	\begin{figure}
		\centering
		\includegraphics[width=0.9\textwidth]{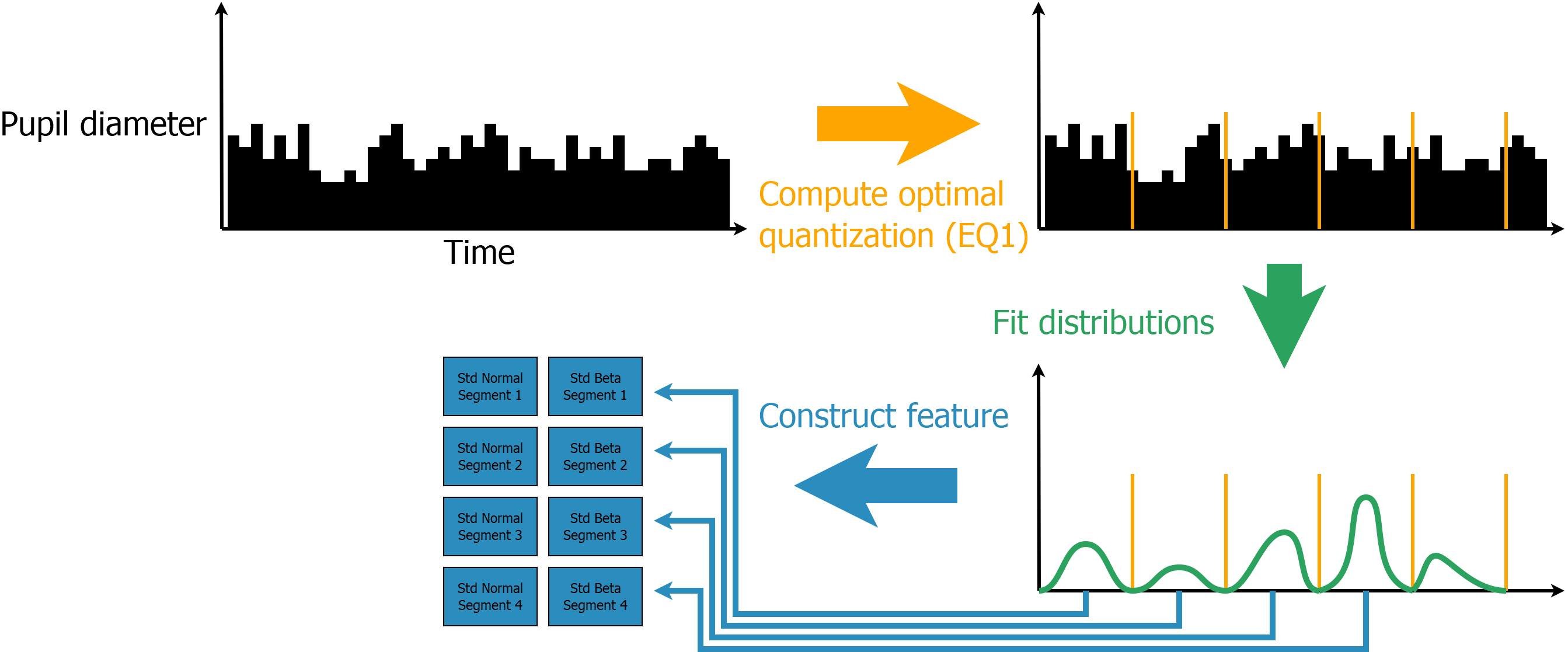}
		\caption{The feature construction workflow. First, we compute an optimal set of splits in the time domain (Orange bars in the top left part, computed with Equation~\ref{eq:EstiSplits}) on the pupil diameters (Black bars in the top left and right part). For each time slot, a Normal and a Beta distribution are fitted (Green line in the bottom right part). The standard deviation of all Normal and Beta distributions are concatenated (Bottom left part) which represents our feature.}
		\label{fig:Workflow}
	\end{figure}
	
\begin{figure}
		\centering
		\includegraphics[width=0.32\textwidth]{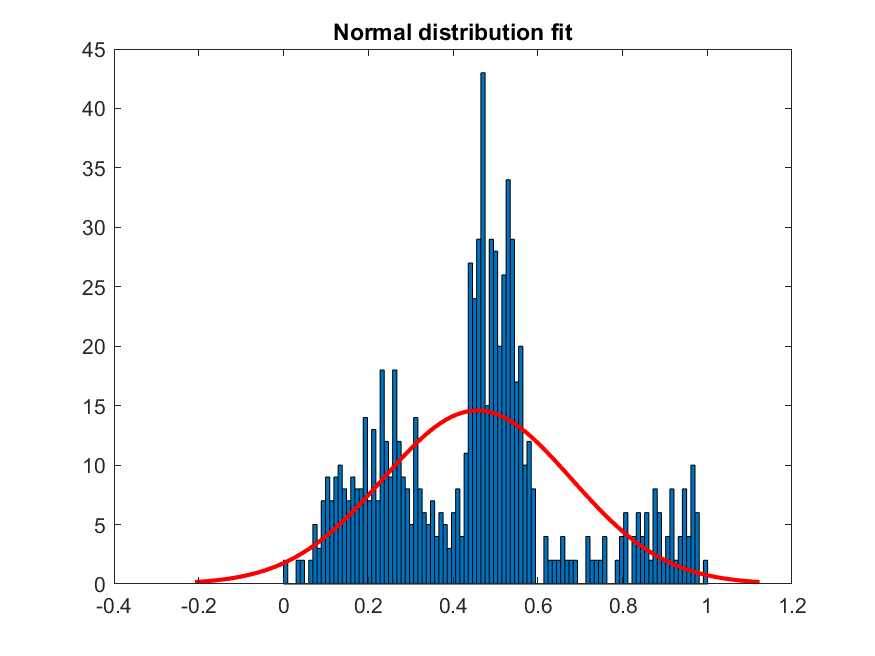}
		\includegraphics[width=0.32\textwidth]{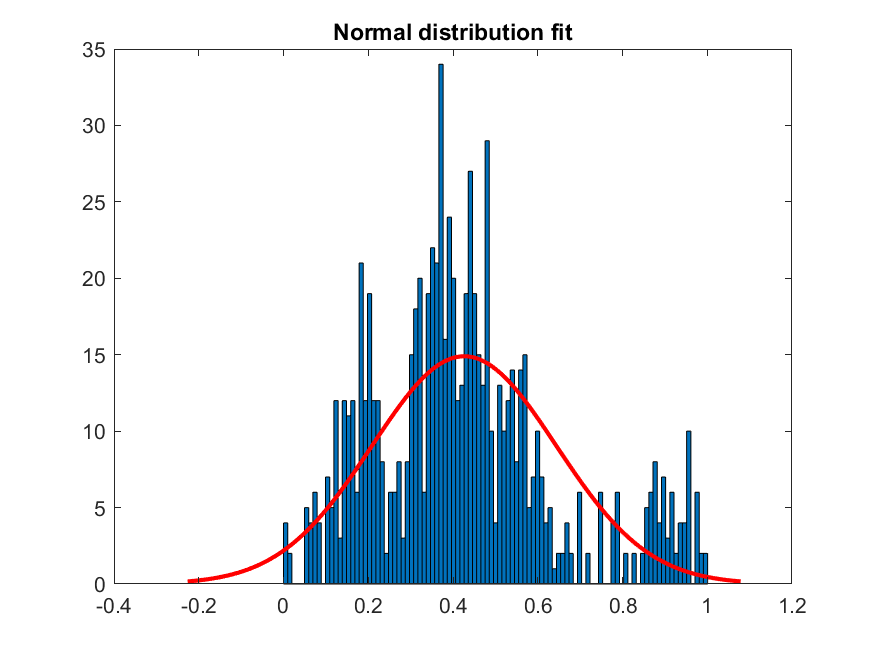}
		\includegraphics[width=0.32\textwidth]{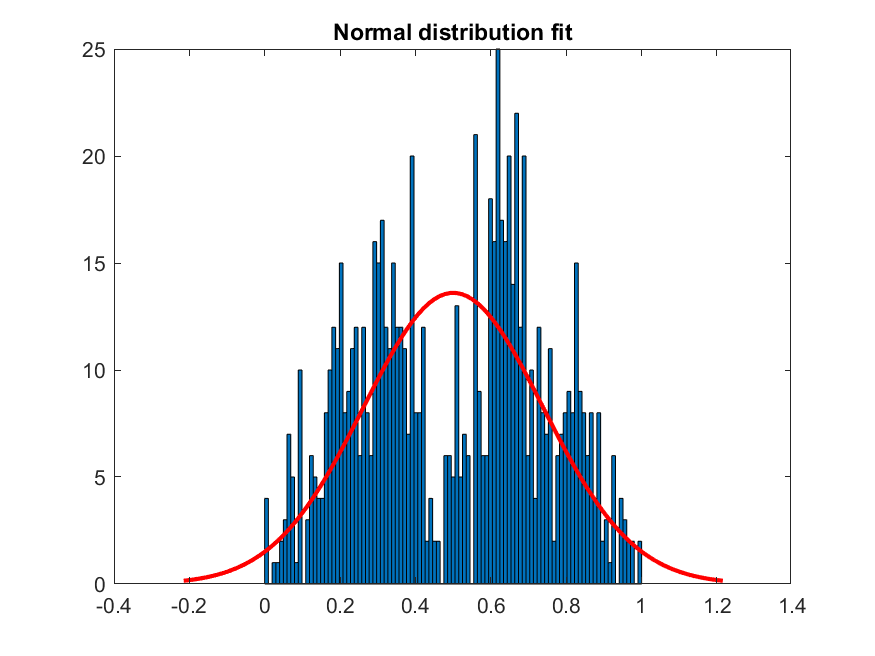}
		\includegraphics[width=0.32\textwidth]{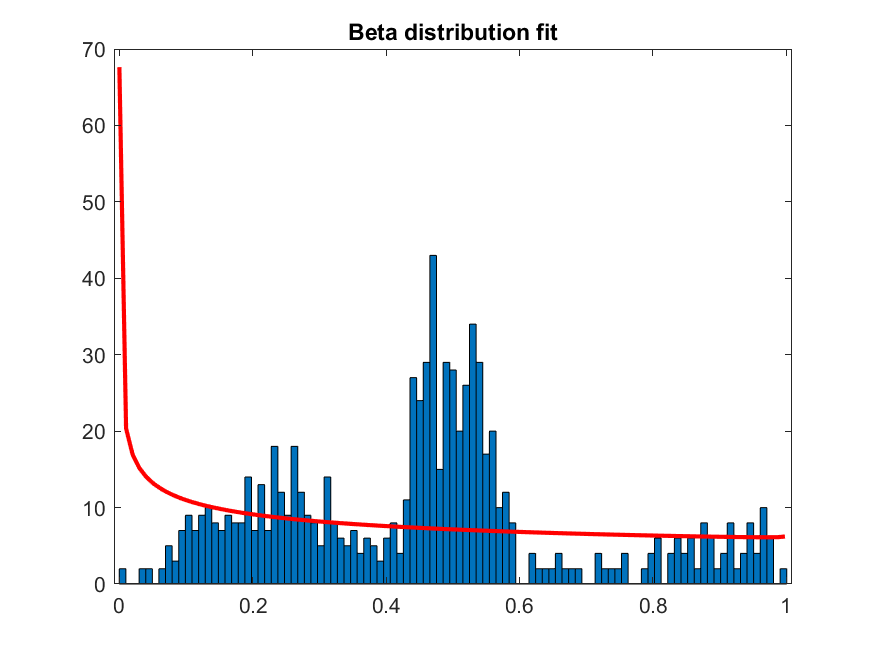}
		\includegraphics[width=0.32\textwidth]{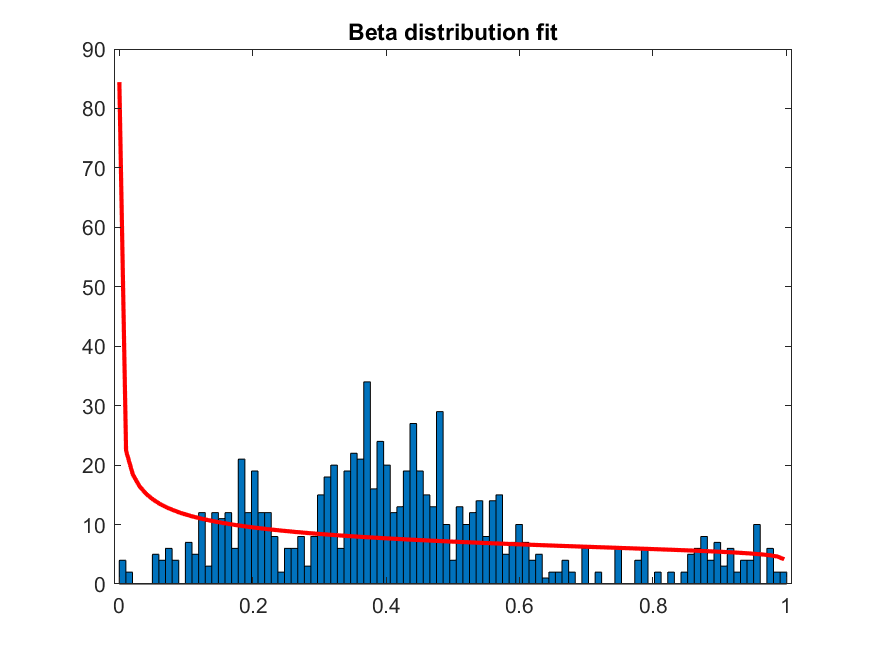}
		\includegraphics[width=0.32\textwidth]{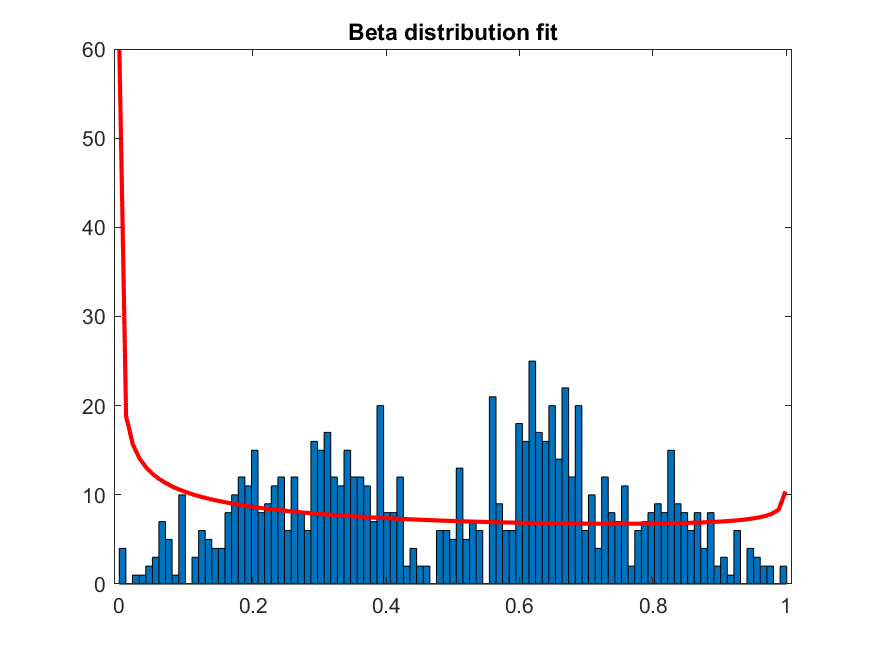}
		\caption{Real distribution fits of a recording of which we use the parameters of the distributions as feature. The top images show the Normal distribution fit (Red) and the real values (Blue). In the bottom images, the Beta distribution fits (Red) are shown with the real values in blue. \textbf{The data of two superimposed images is identical, but the range of values for the normal distribution is -0.4 to 1.2 or -0.2 to 1.2 (Last image) to show the variance of the distribution. This makes the data look compressed compared to the beta distribution, which is defined only in the range 0 to 1.} Each superimposed pair of images represents one time based quantization index, and therefore this figure shows three time slots.}
		\label{fig:Feature}
	\end{figure}

	Figure~\ref{fig:Workflow} shows the process of our feature construction. We only use the pupil diameter for our feature since it is proven that the pupil dialation contains information about cognitive load~\cite{beatty1982task,CORR2016FuhlW,WDTE092016,WTCDAHKSE122016,zekveld2011cognitive}. The pupil diameters are shown as black bars over the time (X axis) in Figure~\ref{fig:Workflow}. In the first step, we calculate an optimal division of time into equal intervals. This is done only on a part of the training data and described together with Equation~\ref{eq:EstiSplits} in detail at the end of this section. The idea behind the split into time intervals is, that it contains more fine grained information in comparison to break the pupil dialation in total down to two or three statistical values. In addition, it makes it possible for the machine learning approach to compare the values from different time segments. After the split into time segments, we fit a Normal and a Beta distribution to each segment. From both distributions we only use the standard deviation since we are only interested in the pupil dialation itself and not on the actual pupil size. Before the fits, we also normalize the pupil diameters per time segment into the range 0 to 1, this should remove effects like people having a naturally larger pupil size and therefore, aligne the pupil sizes between the subjects. We decided to use the Normal distribution since in nature nearly everything should behave like the Normal distribution. In addition, we used the Beta distribution since other effects like the eye tracker inaccuracy will have an influence on the pupil diameters. Fits on real data can be seen in Figure~\ref{fig:Feature}.
	
	Figure~\ref{fig:Feature} shows the structure of our feature with real data. In Figure~\ref{fig:Feature} all blue entries represent the original data of one recording. The red curves represent fitted probability distributions. The top plots show the fitted Normal distribution in red and the bottom images the fitted Beta distributions also in red. Two superimposed images belong to the same data. This means that the blue bars in the histogram are identical in both images. For the Normal distribution it looks like the blue bars are compressed, but this is due to a different value range (X axis) which is stretched to show the Normal distribution correctly. The six plots in Figure~\ref{fig:Feature} correspond to three time slots. Those time slots are splits of the recording into time areas. Each pair of superimposed images corresponds to the same time slot or split index. 
	
	\begin{equation}
		\label{eq:EstiSplits}
		\argmin_{Splits} = \sum_{i=1}^{Splits} |N_i-R_i|+|B_i-R_i|
	\end{equation}
	
	Equation~\ref{eq:EstiSplits} describes the process to split the data in the time domain. In general, we always split the data uniformly, which means that each block has the same size over the time domain, but the amount of splits is determined by Equation~\ref{eq:EstiSplits}. In Equation~\ref{eq:EstiSplits} $Splits$ is the amount of splits over the time domain, $N_i$ is the fitted Normal distribution, $B_i$ is the fitted Beta distribution, and $R_i$ is the real distribution as histogram. With $|N_i-R_i|$ we receive the absolute difference between the Normal and the Real distribution, and with $|B_i-R_i|$ the absolute difference between the Beta and the Real distribution. Summing over all splits, we search for the amount of Splits which minimized the difference for all distribution fits. In practice, we select a random set of samples from the training set and find for each sample the optimal fit. Afterwards, the mean over all found Splits is selected. We use a random subset here to avoid overfitting to the training set.
	
	So far, we only described our approach in a way to apply it to an entire recording, but our approach is also online applicable. For this, we do not use the entire recording, instead we split the recording into window sizes based on the time and compute our feature for each window separately. This way we can use the proposed feature in an online fashion. We also evaluated the proposed approach this way with different window sizes, as can be seen in Table~\ref{tbl:ONLINE}.

	\section{Evaluation}
	In this section, we compare our feature to other SOTA features from \cite{ktistakis2022colet}. In the first part, we describe the used public data set. Afterwards, we describe our hardware setup, the used software, as well as the used parameters for the machine learning approaches. All evaluations are based on the NASA-TLX scores~\cite{hart1988development}, which is the mostly used metric in cognitive load assessment.
	
	We used the data set provided in \cite{ktistakis2022colet} with 47 subjects. In total the authors recorded 56 subjects but 9 got excluded due to poor quality recordings (2) and ocular disease (7). From these 47 subjects, 26 are female and 21 are male. The mean age of the subjects was 32 with a standard deviation of 8 years. Education level of the subjects was 17 years in average with a standard deviation of 2 years. In total, each subject had to perform four activities, which resulted in 188 recordings. The four activities are divided in no time pressure and single task, time pressure and single task, no time pressure and multitasking, and time pressure and multitasking. The main task was to indicate one of nine squares with a certain object. For the secondary task, the subjects had to count aloud and backwards from 1000 by subtracting 4. For each activity and participant, 5 images were selected randomly from the image pool (21 images, each split into 9 boxes). The activities were also presented in a randomized order to avoid any learning and fatigue effects. After each activity, the subjects had to fill out the NASA-TLX questionnaire~\cite{hart1988development}. Based on this questionnaire the Mental Demand, Physical Demand, Temporal Demand, Performance, Effort, and Frustration are acquired. For each recording, the mean of these six scores is computed and divided into three classes. These classes are $C1=Mean~NASA~TLX~score~<~30$, $C2=Mean~NASA~TLX~score~30~<~C2~<~50$, and $C3=Mean~NASA~TLX~score~>~50$. For the recordings, the authors used the Pupil Core eye-tracker from Pupil Labs~\cite{kassner2014pupil} with 240 Hz sampling frequency. During the recording, they used a chin and head rest to minimize head movements. Each subject was seated at 80 cm distance to the screen (24 inch LCD display with $1280 \times 720$ pixels resolution). All recordings are performed under controlled lighting conditions (Screen off 400 lx and blank screen 450 lx).
	
	For our evaluations, we used Matlab 2022b on a Windows 10 64-Bit PC with 64 GB DDR4 memory and an AMD Ryzen 9 3950X 16-Core Processor with 3.50 GHz. The used machine learning methods are Random Forrest Ensemble (RF), Gaussian Naive Bays (GNB), Linear Regression (LR), Linear Support Vector Machine (SVM), K Nearest Neighbors (KNN), Discriminant Analysis (DR), and Neural Networks (NN). For all approaches, we used the default parameters in Matlab 2022b. The only parameter we set manually was the amount of random forests in the ensemble to train, which we set to 10. For the evaluation we used the metrics Accuracy ($\frac{TP+TN}{TP+TN+FP+FN}$), Precision ($Mean(\frac{TP}{TP+FP})$), Recall ($Mean(\frac{TP}{TP+FN})$), and F1 Score ($Mean(\frac{2TP}{2TP+FP+FN})$) for the classification task and the mean absolute error for regression. With TP we mean true positives, with TN the true negatives, with FP the false positives, and with FN the false negatives. The $Mean()$ in the Precision, Recall, and F1 Score donates the averaging over each class. This means we computed the Precision for each class seperately and averaged the result which is exactly the definition of the metrics for a multiclass problem. 
	
	The SOTA features are fixation frequency, fixation duration, fixation duration variation, saccade frequency, saccade amplitude variation, saccade velocity, saccade velocity variation, saccade velocity skewness, saccade velocity kurtosis, peak saccade velocity, peak saccade velocity variation, peak saccade velocity skewness, peak saccade velocity kurtosis, saccade duration, saccade duration variation, blink frequency, blink duration, pupil diameter, pupil diameter variation, pupil diameter skewness, and pupil diameter kurtosis as described in \cite{ktistakis2022colet}. Each feature was normalized to the range 0 to 1 as it was done by the authors in \cite{ktistakis2022colet}. For the train test split, we also used the same split as in \cite{ktistakis2022colet}, which is a random split using 80\% of the data for training as well as validation and the other 20\% for testing. For the estimation of the amount of splits for our feature, we used 30\% of the training data. With those 30\% we got a split size of 10 for the experiments with the entire sequence ((Table~\ref{tbl:VSSOTA}) and (Table~\ref{tbl:REG})) and 5 for the online experiment with different window sizes (Table~\ref{tbl:ONLINE}).

	\begin{table}
		\centering
		\caption{Classification results in comparison to the SOTA features for the three classes ($C1=Mean~NASA~TLX~score~<~30$, $C2=Mean~NASA~TLX~score~30~<~C2~<~50$, $C3=Mean~NASA~TLX~score~>~50$). We evaluated different machine learning approaches namely Random Forrest (RF), Gaussian Naive Bays (GNB), Linear Regression (LR), Linear Support Vector Machine (SVM), K Nearest Neighbors (KNN), and Discriminant (DR). The used metrics are the accuracy, precision, recall, and the f1 score.}
		\label{tbl:VSSOTA}
		\begin{tabular}{cccccc}
			Feature  & ML Method & Accuracy & Precision & Recall & F1 \\ 
			\multirow{6}{*}{\cite{ktistakis2022colet}}  & RF & 60.53 & 58.51 & 57.08 & 57.18 \\ 
			& GNB & 57.89 & 45.79 & 46.83 & 42.75 \\ 
			& LR & 55.26 & 43.22 & 44.44 & 39.07 \\ 
			& SVM & 60.53 & 46.15 & 39.15 & 41.80 \\ 
			& KNN & 57.89 & 56.65 & 55.87 & 55.68 \\ 
			& DR & 52.63 & 40.60 & 44.23 & 36.06 \\   \hline
			\multirow{6}{*}{Proposed}  & RF & \textbf{68.42} & \textbf{65.49} & \textbf{65.49} & \textbf{65.49} \\ 
			& GNB & 52.63 & 39.89 & 33.33 & 36.06 \\ 
			& LR & 65.79 & 59.30 & 59.68 & 59.03 \\ 
			& SVM & 65.79 & 62.21 & 61.67 & 61.48 \\ 
			& KNN & 55.26 & 51.89 & 51.50 & 51.41 \\ 
			& DR & 57.89 & 55.23 & 52.71 & 53.03 \\   \hline
		\end{tabular}
	\end{table}
	
	Table~\ref{tbl:VSSOTA} shows the results for the proposed feature with different machine learning approaches in comparison to the SOTA features from \cite{ktistakis2022colet}. We evaluated the features from \cite{ktistakis2022colet} again to have a fair comparison, since the random 80\% to 20\% split may differ based on the random number initialization. We nearly achieved the same result for GNB (Gaussian naive bays) and an even better result for the random forest (RF) as well as the linear support vector machine (SVM). In comparison to the proposed feature, the SOTA features are outperformed for nearly all machine learning methods. The only machine learning approach where the proposed feature is outperformed is the Gaussian naive bays (GNB) classifier. Comparing all metrics and all machine learning approaches, the best result is achieved by the random forest (RF) with the proposed feature. The second-best result is achieved by the support vector machine (SVM) with the proposed feature, and the third best is the linear regression (LR) with the proposed feature. Therefore, we argue that the proposed feature outperforms the state-of-the-art.
	
	\begin{table}
		\centering
		\caption{Classification results for different window sizes in comparison to the SOTA features for the three classes ($C1=Mean~NASA~TLX~score~<~30$, $C2=Mean~NASA~TLX~score~30~<~C2~<~50$, $C3=Mean~NASA~TLX~score~>~50$). The window sizes are given in seconds, which can be directly translated to the sample size by multiplying them with the  eye tracker camera sampling rate (240Hz). We evaluated two machine learning methods, namely Random Forrest (RF) and Linear Support Vector Machine (SVM). The used metrics are the accuracy, precision, recall, and the f1 score. Recordings with a lower sample size than two times the window size are excluded from the evaluation for the specific window size.}
		\label{tbl:ONLINE}
		\begin{tabular}{ccccccc}
			Feature  & ML Method & Window (Sec.) & Accuracy & Precision & Recall & F1 \\ 
			\multirow{5}{*}{\cite{ktistakis2022colet}} & \multirow{5}{*}{RF} & 10 & 47.37 & 42.00 & 41.69 & 41.81 \\ 
			& & 15 & 55.26 & 52.73 & 50.97 & 50.95 \\ 
			& & 20 & 47.37 & 61.21 & 60.91 & 59.63 \\ 
			& & 25 & 63.16 & 60.48 & 62.33 & 61.04 \\ 
			& & 30 & 63.16 & 66.21 & 65.56 & 64.43 \\    \hline
			\multirow{5}{*}{Proposed} & \multirow{5}{*}{RF}& 10 & 65.79 & 62.21 & 61.67 & 61.48 \\ 
			& & 15 & 60.53 & 55.24 & 56.88 & 55.51 \\ 
			& & 20 & 68.42 & \textbf{77.58} & \textbf{76.50} & \textbf{75.48} \\ 
			& & 25 & \textbf{71.05} & 68.95 & 71.80 & 69.06 \\ 
			& & 30 & 65.79 & 68.03 & 70.29 &  68.30\\ \hline \hline
			\multirow{5}{*}{\cite{ktistakis2022colet}} & \multirow{5}{*}{SVM} & 10 & 55.26 & 42.45 & 46.02 & 37.59 \\ 
			& & 15 & 55.26 & 43.22 & 35.22 & 38.74 \\ 
			& & 20 & 36.84 & 54.55 & 42.38 & 46.75 \\ 
			& & 25 & 50.00 & 49.12 & 50.97 & 46.07 \\ 
			& & 30 & 57.89 & 60.10 & 58.33 & 55.00 \\   \hline
			\multirow{5}{*}{Proposed} & \multirow{5}{*}{SVM}& 10 & 50.00 & 43.14 & 44.36 & 41.62 \\ 
			& & 15 & 50.00 & 41.99 & 43.89 & 42.29 \\ 
			& & 20 & 57.89 & \textbf{68.48} & 66.67 & 66.78 \\ 
			& & 25 & \textbf{71.05} & 65.51 & 66.11 & 65.59 \\ 
			& & 30 & 65.79 & 67.47 & \textbf{68.59} & \textbf{67.94} \\  \hline
		\end{tabular}
	\end{table}

	So far we used entire recordings to compute the features. This does not work in an online case, where a scientist wants to get the information after a certain time regularly. Therefore, we designed another experiment where all the recordings are split into subsequences with fixed window sizes based on the time domain. For the splitting into window sizes, all recordings smaller than two times the current window size are removed from the evaluation. All other recordings are split up based on the window size, with a step size of $\frac{window~size}{2}$. This was done to avoid very large recordings to have a huge impact, since large recordings with a step size of one would produce a multiple of samples compared to small recordings. As machine learning approaches, we selected the support vector machine (SVM) and the random forest (RF) since the SOTA features worked best with those two methods. The results can be seen in Table~\ref{tbl:ONLINE}. For the support vector machine (SVM) the SOTA features outperform the proposed feature up to a window size of 15 seconds. Afterwards, they are outperformed by the proposed feature. In addition, the proposed feature with a window size of 25 seconds outperforms the experiment where the entire recordings are used (Table~\ref{tbl:VSSOTA}). Our explanation for this is that the size alignment of the recordings has a positive effect on the result and can be understood as a normalization step.
	
	For the random forest (RF), the SOTA feature is outperformed for all window sizes in all metrics by the proposed feature. Comparing the results of the window size 20 and 25 for the proposed feature with the results from Table~\ref{tbl:VSSOTA} it is obvious that the size alignment has a huge and positive effect on the results. In terms of accuracy we achieve the same result with a window size of 20, but the other metrics (Precision, Recall, F1 score) show that the classifier with window size 20 is superior. For the window size of 25 all metrics show that the classifier trained on the entire recordings is outperformed.
	
	\begin{table}
		\centering
		\caption{Regression results for the mean NASA TLX scores which consist of Mental Demand, Physical Demand, Temporal Demand, Performance, Effort, and Frustration. We computed the mean absolute error (MAE) on the test set as a comparison metric. We used neural networks (NN) as machine learning approach with one hidden layer. The size of the hidden layer is indicated as the number after the hyphen.}
		\label{tbl:REG}
		\begin{tabular}{ccc}
			Feature  & ML Method & MAE  \\ 
			\multirow{3}{*}{\cite{ktistakis2022colet}}  & NN-5 & 15.55  \\ 
			& NN-10 & 15.16 \\ 
			& NN-20 & 15.32 \\ \hline
			\multirow{3}{*}{Proposed}  & NN-5 & \textbf{12.85} \\ 
			& NN-10 &  13.04\\ 
			& NN-20 & 13.18 \\  \hline
		\end{tabular}
	\end{table}
	
	Table~\ref{tbl:REG} shows the results of our regression experiment with the metric mean absolute error (MAE). In this experiment, we tried to regress the mean NASA TLX score directly instead of splitting the score up into three classes. The mean NASA TLX score is computed as the mean from the Mental Demand, Physical Demand, Temporal Demand, Performance, Effort, and Frustration scores of the NASA TLX questioner. In Table~\ref{tbl:REG} it can be seen that with the machine learning method neural network (NN) the proposed feature outperforms the SOTA feature for all configurations of the neural network. The configuration of the network is noted as a number after the hyphen, which specifies the size of the hidden layer of the two layered neural network. For the proposed feature, the best configuration was the two layered neural network with a hidden layer size of 5 and for the SOTA feature the neural network with hidden layer size of 10. Comparing the best configuration for both, the proposed feature improves the result by 2,31 scoring points as average over all results.
	
	\section{Limitations}
	Our approach outperformed the state-of-the-art, but still there are some limiting factors behind this research. First, the data we used for our evaluation are recorded under laboratory conditions. This means, that there were controlled light conditions and the user could not freely move. For a real world everyday live application, those are hard perquisites, especially the changing lightning conditions or the flickering from fluorescent tubes. In contrast to this limitation, our feature can still help science in experiments and studies. Additionally, it is also applicable in areas where the lightning conditions can be controlled, like medical measurements for a diagnosis.
	
	Another limiting factor of our approach is that the results are not perfect. This means, that there is still room for a failure or a false classification. Therefore, the results in applications must be taken with care and there should be a mechanism to estimate the uncertainty in the results.
	
	\section{Conclusion}
	In this paper, we proposed a novel feature for cognitive load classification and mean NASA TLX score regression. The feature itself has no spatial dependent features contained and is only based on the pupil variations in the data itself. Therefore, the feature does not contain risks of experimental influences like fixations and saccades do, since those are influenced by the task as well as the stimulus itself. In addition, we showed that our feature is online capable by using a time window. In all experiments our proposed feature outperforms the SOTA features under four metrics (Accuracy, Precision, Recall, F1 score). Therefore, we argue that the proposed feature is a valuable contribution for the research community. We hope that our feature helps to bring cognitive load classification into real world applications and supports scientists in their evaluations.

	\bibliographystyle{plain}
	\bibliography{template}

\end{document}